\documentclass[conference]{IEEEtran}
\IEEEoverridecommandlockouts
\usepackage{cite}
\usepackage{amsmath,amssymb,amsfonts}
\usepackage{algorithmic}
\usepackage{graphicx}
\usepackage{textcomp}
\usepackage{xcolor}
\usepackage{pgfplots}
\def\BibTeX{{\rm B\kern-.05em{\sc i\kern-.025em b}\kern-.08em
    T\kern-.1667em\lower.7ex\hbox{E}\kern-.125emX}}
\begin{document}

\title{{Science based AI model certification for new operational environments with application in traffic state estimation}\\
}

\author{\IEEEauthorblockN{Daryl Mupupuni\IEEEauthorrefmark{1},
Anupama Guntu\IEEEauthorrefmark{2}, Liang Hong\IEEEauthorrefmark{3},
Kamrul Hasan\IEEEauthorrefmark{4} and
Leehyun Keel\IEEEauthorrefmark{5}}
\IEEEauthorblockA{Department of Electrical and Computer Engineering,
Tennessee State University\\
Nashville, TN 37209, USA\\
Email: \IEEEauthorrefmark{1}darylmupupuni@gmail.com,
\IEEEauthorrefmark{2}anupamaguntu23@gmail.com,
\IEEEauthorrefmark{3}lhong@tnstate.edu, \\
\IEEEauthorrefmark{4}mhasan1@tnstate.edu,
\IEEEauthorrefmark{5}keel.tsu@gmail.com}}

\maketitle

\begin{abstract}
The expanding role of Artificial Intelligence (AI) in diverse engineering domains highlights the challenges associated with deploying AI models in new operational environments, involving substantial investments in data collection and model training. Rapid application of AI necessitates evaluating the feasibility of utilizing pre-trained models in unobserved operational settings with minimal or no additional data. However, interpreting the opaque nature of AI's black-box models remains a persistent challenge. Addressing this issue, this paper proposes a science-based certification methodology to assess the viability of employing pre-trained data-driven models in new operational environments. The methodology advocates a profound integration of domain knowledge, leveraging theoretical and analytical models from physics and related disciplines, with data-driven AI models. This novel approach introduces tools to facilitate the development of secure engineering systems, providing decision-makers with confidence in the trustworthiness and safety of AI-based models across diverse environments characterized by limited training data and dynamic, uncertain conditions. The paper demonstrates the efficacy of this methodology in real-world safety-critical scenarios, particularly in the context of traffic state estimation. Through simulation results, the study illustrates how the proposed methodology efficiently quantifies physical inconsistencies exhibited by pre-trained AI models. By utilizing analytical models, the methodology offers a means to gauge the applicability of pre-trained AI models in new operational environments. This research contributes to advancing the understanding and deployment of AI models, offering a robust certification framework that enhances confidence in their reliability and safety across a spectrum of operational conditions.
\end{abstract}

\begin{IEEEkeywords}
Artificial Intelligence (AI), Model Certification, Scientific Knowledge, Physical Inconsistency, Traffic State Estimation (TSE)
\end{IEEEkeywords}

\section{Introduction}
Artificial intelligence (AI) has experienced tremendous growth and application across several fields \cite{amin2024explainable,amin2022industrial,amin2022analysing}. Application of AI has completely changed the engineering sector over the last ten years	by allowing engineers to design creative ideas, solve complex problems, and expedite operations in several engineering disciplines. For instance, various environmental engineering problems were solved by AI based prediction models such as pollution control, wastewater to improve the performance of several chemical and logical treatment processes \cite{Yetilmezsoy2011ARTIFICIALIP,YE2020134279}. AI is also useful in autonomous systems, proactive maintenance in the aerospace and automobile industries \cite{Zhang2017CurrentTI}. AI is changing engineering by automating tasks, improving designs, and enabling predictive analysis. The integration of AI will continue driving innovation in the future \cite{xing2020driving}.

Although AI models like Artificial Neural Networks, Ensemble Approaches etc., continue to perform better than other models in many fields, but their acceptance in delicate fields like finance and healthcare is questionable because of the models’ difficulty in being understood and explained \cite{islam2019infusing}. The lack of interpretability or transparency also known as "black-box models," is a key cause of concern in some AI models and it is challenging to comprehend how they came to those conclusions \cite{zhou2017machine}. The minimal amount of data available for training AI models is another restriction. For AI models to learn effectively, large and varied datasets are necessary to train the models. Owing to the rarity of some events or the biases present in the data, it can be difficult to obtain balanced datasets in some areas. As a result, the model’s predictions may be biased or have skewed representation. This is also reflected in the application of models developed from these datasets to be employed in new operational situations \cite{7906512}.

Since AI models operate as black boxes, they must undergo certification to ensure their effective performance across various environments.  This certification process can take various forms, including assessments that verify adherence to minimum ethical standards and what we term as nuanced evaluations \cite{article}. Some important applications like robots in work spaces shared with humans \cite{winter2021trusted}, social and environmental \cite{gupta2020secure} need to be certified. This type of certification study is also important in traffic state estimation (TSE) which is not fully examined yet. The goal of the certification process is to ascertain whether the TSE model developed under different traffic state conditions can accurately anticipate in a new setting while upholding the traffic conservation rule. By certifying the model, we hope to increase its interpretability, strengthen its dependability, and demonstrate its suitability for TSE example that occur in the real world \cite{tambon2022certify}. The findings of this study have implications for developing the TSE area and advance knowledge of certifying deep learning models for safety-critical applications. 

New operational environments refer to real-world settings where an AI system is deployed that differ from those used to develop and test the system, potentially exposing it to unfamiliar situations and inputs. In this particular instance, the new environments  are new traffic contexts. To fulfill the urgent demand for safety, dependability, and transparency in AI applications, deep learning models need certification \cite{winter2021trusted}. The contributions of this paper are the establishment of a thorough science-based AI model certification methodology that ensures the resilience of deep learning models in new operational situations by thorough examination and the incorporation of known scientific concepts. This work enables one to ascertain the extent to which AI models adapt to new complex and dynamic environments, reducing risks and potential accidents by bridging the gap between conventional data-driven approaches and science-based understanding. Additionally, the focus on transparency and explainability encourages a better understanding of model judgments, fostering public confidence and facilitating regulatory compliance \cite{bakirtzis2022dynamic},\cite{zhang2023ethics}. We seek to increase the precision and dependability of TSE models, resulting in safer and more effective transportation systems.

\subsection{Literature Review }

Certifying AI models is pivotal when deploying them in unfamiliar operational contexts, particularly where conventional data-driven methods may prove inadequate. This is of paramount importance in applications like autonomous vehicles, where certification ensures safety and dependability in diverse and unpredictable conditions such as bad weather and unusual traffic patterns. Similarly, in healthcare diagnostics, certification is necessary to handle different patient populations, rare diseases, and unforeseen medical situations, instilling confidence in the accuracy and clarity of AI-assisted diagnoses especially in new operational environments \cite{azzam2004route}, \cite{yanisky2019equality}. 

In natural disaster prediction and response artificial intelligence models are used to anticipate natural disasters, such as earthquakes, hurricanes, or wildfires, need to be certified to function well in disaster scenarios that haven’t been encountered before \cite{kuglitsch2022facilitating}. These model certifications guarantees their ability to adjust to shifting circumstances and deliver accurate early warnings. In aerospace and aviation AI certification is needed to address the difficulties of new operational contexts, such as air traffic changes, weather disturbances, and emergency scenarios, certification is required for autonomous drones and aircraft systems \cite{bakirtzis2022dynamic}.

In the fields of industrial robotics and manufacturing, artificial intelligence (AI) models must be certified to operate in dynamic production environments with the possibility of varying raw material supply, equipment breakdowns, or unforeseen production difficulties.AI models used for environmental monitoring, such as those that predict weather, ocean currents, or air quality, must be certified to handle a variety of geographic and climatic conditions \cite{SOORI202354}. 

Certifying AI models is paramount in security and surveillance for operation in uncontrolled environments, such as public spaces where unforeseen occurrences and abnormalities may occur \cite{berghoff2020vulnerabilities}. Financial risk assessment also needs certification owing to financial markets being subject to abrupt volatility and unforeseen events, AI models for risk assessment and prediction in finance must be approved for new operational settings. For Power Grid Management, AI models used to manage power grids and energy distribution systems must be certified to react to unforeseen changes in energy supply, demand, and potential grid failures \cite{syu2023ai}. For application in space exploration, AI models used in space missions need to be certified to withstand the harsh environments of uncharted space, where data may be sporadic and unexpected \cite{budnik2018guided}. 

Certification is necessary in order to check for several challenges that are encountered when we are trying to employ pre-trained AI models in new environments. Some of the challenges that may be encountered include models that may be vulnerable to adversarial assaults, in which malicious inputs are skillfully created to falsely estimate outcomes \cite{berghoff2020vulnerabilities}. When pre-trained models are used in new environments, privacy concerns may arise because vulnerable data may unintentionally leak through model outputs. Ethics issues may also arise if AI models are used in new environments without being properly vetted or knowing their limitations, especially if important choices are based on their outputs. Pre-trained models can be used in situations where they are not in compliance with rules or legal requirements, which can result in legal issues and liability issues \cite{hussain2022shape}. Understanding how pre-trained models work can be tricky because they are complex sometimes and difficult to interpret. 

Another important application is traffic state estimation where this certification is mandatory to test the models in different environments.  It is always difficult to acquire huge number of data in all environments since it involves more cost to collect it. In that scenario, certification of AI model in other environments is necessary.  Thus, it is essential to carry out thorough testing, validation, and fine-tuning of the pre-trained model in the particular operating context in order to assure its safety, dependability, and ethical use in order to reduce these risks. Additionally, keeping an eye on the model’s performance and user input can aid in spotting and resolving possible problems \cite{berghoff2020vulnerabilities}.

Pre-trained models can be quite effective and helpful in terms of time and computational power, but they may not always function at their best in novel and uncharted circumstances pausing risks of increased uncertainties. Pre-trained models that are typically trained on a small number of datasets may not have a thorough understanding of domains that are unrelated to their training data. The availability of unbalanced datasets is a frequent problem in traffic state estimation. A dataset that is uneven in the context of the traffic state estimation example would be one where some traffic states or situations are over represented while others are underrepresented or rare \cite{liang2018deep}. The imbalance may be caused numerous things, including traffic patterns, variable levels of congestion, or the scarcity of data from particular places or times. For instance the traffic dataset that may be readily available is mainly the NGSIM-CA data from California and the traffic state may not fully represent the state of traffic in other states say Tennessee.  The model may generate inaccurate or deceptive results in new circumstances as a result of its ignorance of the relevant domain. In some cases the input data in a context for which the model has not yet been trained differs from that context, the model may interpret the input incorrectly and produce false results.

Pre-trained models may inherit biases from their training data, which is known as bias amplification. These biases may intensify when applied in unknown environments, producing unfair or discriminating outcomes. It is possible for pre-trained models to over-fit to specific applications or datasets \cite{zhou2020sentix}. They might not generalize when applied in novel and varied circumstances.  Due to domain variations, the model's performance may suffer in new circumstances, producing inconsistent results and decreasing user confidence.

It can be very beneficial to use pre-trained AI models in new environments, but doing so requires careful preparation, ongoing monitoring, and the right tweaks to make sure the model operates properly and safely. Striking a balance between using the model's pre-trained skills and tailoring it to the unique demands of the operational setting is crucial.

\subsection{Methodology}

This section of the paper provides background information on traffic flow physics and its relevance to AI models, particularly in the context of traffic data certification. It emphasizes the importance of understanding fundamental traffic concepts, illustrated by the fundamental traffic diagram (Figure 1), which depicts the linear relationship between traffic speed and density \cite{wang2020estimating,RAISSI2019686,lv2014traffic}. The conservation of vehicles principle, based on mass conservation, is introduced as a fundamental physics concept governing traffic flow. Additionally, mathematical models such as the Lighthill-Whitham-Richards (LWR) model, based on fluid dynamics principles, are discussed for representing traffic flow and analyzing congestion \cite{raissi2018deep,10105558}.

\begin{figure}
	\centering 
	\includegraphics[width=0.5\textwidth, angle=0]{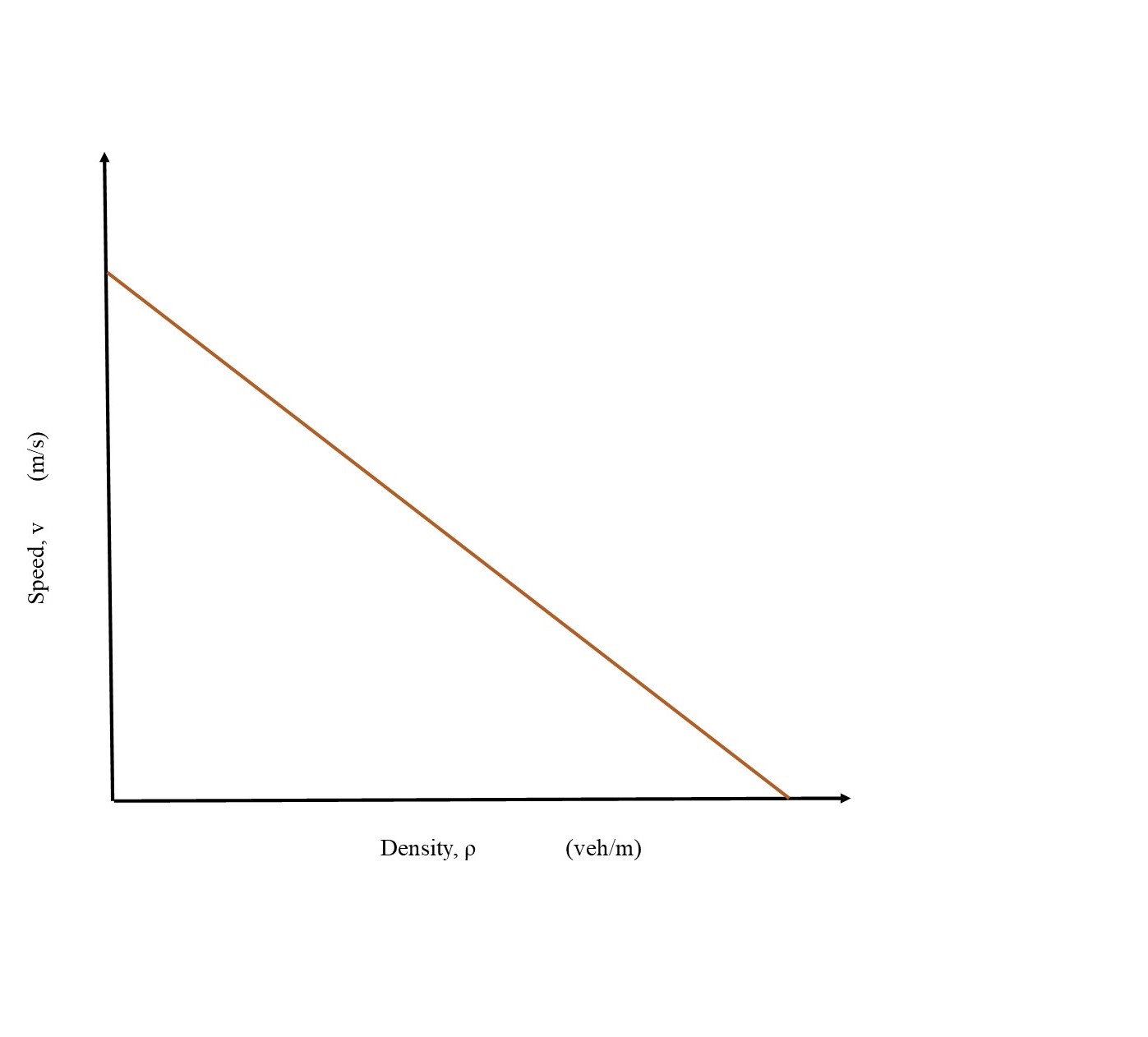}	
	\caption{Speed against density diagram in traffic flow } 
	\label{fig_mom0}%
\end{figure}

The section delves into the relationship between traffic density, flow rate, and vehicle speed, highlighting the impact of these factors on traffic conditions \cite{park2016real}. It also introduces equations describing traffic flow, including the LWR equation. The Greenshields' fundamental diagram is presented to summarize the relationship between traffic density, flow rate, and free-flow speed is mathematically defined by equations 1 and 2 \cite{greenshields1935study,XING2022127079}.

\begin{equation}
q(x,t) = \frac{\partial N(x,t)}{\partial t}
\end{equation}

\begin{equation}
\rho(x,t) = -\frac{\partial N(x,t)}{\partial x}
\end{equation}

The discussion transitions to the application of Artificial Intelligence (AI) in traffic state estimation, categorizing it as a data-driven approach. The potential of AI algorithms, particularly deep learning models, in forecasting traffic conditions is emphasized, along with the importance of labeled training data and computational resources \cite{9744160,8694956}. The structure of deep learning neural networks and the role of the cost function in training are explained. The paper provides equations for the mean squared error cost function used in traffic state estimation for the assessment of the metrics.The cost function, also known as the loss or objective function, is a crucial mathematical measure in deep learning that quantifies the disparity between a neural network's expected output and the actual target values \cite{10105558}.

Accuracy in deep learning is explored, focusing on classification and regression accuracy calculations and is defined by the equation. 

\begin{equation}
    A_{D_L} = \frac{\sqrt{\sum_{x=1}^{X_m}\sum_{t=1}^{T_n}|\rho(x,t)- \hat{\rho}(x,t)|^2}}{\sqrt{\sum_{x=1}^{X_m}\sum_{t=1}^{T_n}|\rho(x,t)|^2}}
\end{equation}

Equation 3 is used to evaluate the performance of an AI model and the results obtained from this equation are used to dteremine the errors associated with the predictions. 

Regression accuracy is specifically discussed in the context of traffic state estimation, introducing the Root Mean Squared Error (RMSE) formula. 

\begin{equation}
    C_{D_L} = \text{MSE}_{(\rho(x,t),\hat{\rho}(x,t))} = \frac{1}{N} \sum_{i=1}^{N} |\rho(x,t) - \hat{\rho}(x,t)|^2
\end{equation}

Equation 4 is used in the training process. N is the number of estimating outputs in Mean Squared Error MSE. The estimated vehicle density $\hat{\rho}(x,t)$ and the actual vehicle density, $\rho(x,t)$ at location x and time t.

\section{Methodology Discussion: Model Assessment against Science Metrics Laws}
To evaluate and certify AI models, the steps shown in figure 2 are followed.

Certifying the suitability of a Traffic State Estimation AI model for deployment in new environments necessitates the integration of scientific principles and traffic laws into the model assessment. This involves a comprehensive examination of the model's predictions in alignment with established scientific principles, with a primary focus on the implementation of the conservation of traffic. This approach serves to foster the development of dependable models that strictly adhere to physical and mathematical constraints, thereby establishing a foundational framework for the scientific validation and acceptance of the AI model. Upon saving the model parameters, a crucial step ensues, where the models undergo comparison with well-known physical constraints and physics laws. This comparative analysis aims to augment the reliability, robustness, and trustworthiness of deep learning models designed for traffic state estimation.

The pivotal role played by the law of conservation of traffic in this process is evident as predictions from the model are scrutinized for deviations from ideality. Ensuring adherence to fundamental principles involves applying traffic flow conservation rules or laws, validating that the predicted values for density, velocity, and predicted flow rate align with the actual influx and outflow of vehicles. The assessment extends to confirming the consistency of predicted values for flow rate and density with the anticipated connections suggested by the traffic flow model. An analysis of results is then conducted to identify anomalies or deviations from the established rules of traffic flow, scrutinizing expected density values and traffic variables. This meticulous examination identifies areas, such as instances of congestion or unexpected correlations between flow rate and density, facilitating a comprehensive evaluation of the model's adaptability to new environments.

\section{Experimental Configuration}
Assessment of the performance of the AI system's correctness and dependability are required for science-based certification of AI models. \cite{tambon2022certify}. This section introduces a science-based certification framework for AI models, emphasizing the evaluation of assessment in new environments. The certification process is outlined in Figure 2.

\begin{figure}
	\centering 
	\includegraphics[width=0.5\textwidth, angle=0]     
        {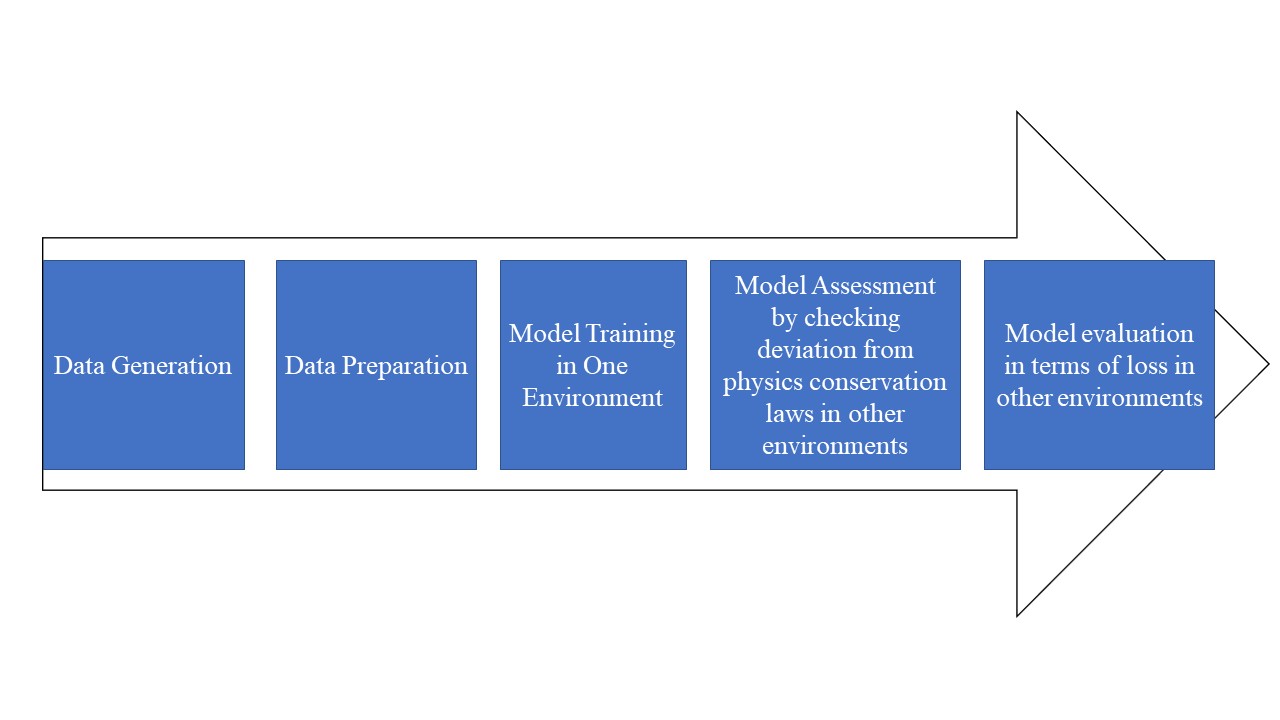}	
	\caption{Certification process} 
	\label{fig_mom0}%
\end{figure}

In this paper, the Lax-Hopf method is utilized for generating synthetic traffic datasets under conditions where there is neither downstream flow nor upstream flow. Moreover, it is specifically calibrated to conform to the Greenshields' model and traffic conservation laws.
\begin{figure}
	\centering 
	\includegraphics[width=0.5\textwidth, angle=0]{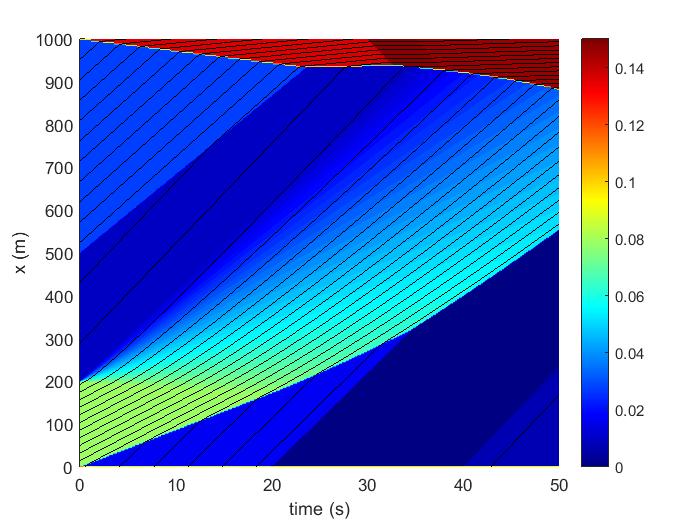}	
	\caption{Dataset generated with a $v_f$ =25} 
	\label{fig_mom0}%
\end{figure}
This experiment is done on a road segment defined by 1000 meters (x) observed over a duration of 50 seconds (t). The traffic density values throughout the domain were created by the Lax-Hopf method by setting up the following initial density values $\rho(x,t)$ and initial flow rates.

Initial density: For interval at x in  [0, 200] meters is 0.13 vehicle per meter (veh/m), [200, 500] is 0.06 veh/m and [500, 1000] is 0.03 veh/m at t=0.

Free flow velocity ($v_f$) of 25 meters per second (55 miles per hour) and jam density ($\rho_m$) of 0.15 veh/m were utilized. The dataset was created at the time steps of 0.1 seconds and the distance step is 2 meters. The generated synthetic dataset is displayed in figure 3.

The deep learning experimental set up uses a 10 hidden layer architecture with 40 neurons each. Two optimization algorithms were used that is the limited Broyden–Fletcher–Goldfarb–Shanno (L-BFGS) and Adam. The Tensorflow library for machine learning was implemented in the development of our Traffic State Estimation.The machine used to run the TSE was a 11th Gen Intel® Core™ i9-11900KF @ 3.50GHz × 16, with a Random Access Memory of 62.5GB. 

\section{Results and Discussions}

This study conducted an insightful investigation into the conservation law of traffic flow. It did so through a two-step modelling approach combined with simulations of varied traffic conditions. First, a machine learning model was trained on sensor data to estimate real-time traffic states like speed and flow. Road geometry and counts then fed into a physics-based conservation formula to calculate expected outflows. By generating new environments with different $v_f$ values while keeping other parameters constant, we aimed to assess the impact on the conservation law. While the conservation law provides a fundamental framework for understanding traffic flow, it is crucial to acknowledge potential deviations in real-world scenarios, influencing traffic modeling and analysis.

The estimated results from the ML model were then directly compared against the calculations derived from the established traffic flow physics model. The differences or errors between the two results were analyzed. This section further underscores the significance of certifying deep learning models with physics laws to enhance predictions in Traffic State Estimation (TSE). The deep learning model, trained in a specific environment and tested across diverse settings, particularly aligns with scenarios featuring sensors strategically positioned within the road network. This approach allows for the observation and analysis of traffic variables at fixed locations over time, mirroring real-world setups with fixed sensors.

The monitored traffic conditions and behaviors at predetermined positions contribute to predicting traffic conditions in new or unobserved areas. Through this comprehensive exploration, we emphasize the robustness of our approach, bridging the gap between deep learning models and the underlying physics governing traffic dynamics for more accurate TSE predictions. The graphical representation in Figure 4 visually reinforces the trends observed, validating the effectiveness of our methodology in real-world applications.

The deep learning (DL) model undergoes training using only 6\% of the complete dataset, specifically 15,000 samples, with a fixed $v_f$ of 25. The testing phase employs the remaining dataset, encompassing data from various environments. During training, we utilized 15,000 iterations with the Adam optimizer and an additional 50,000 iterations with the L-BFGS optimizer. 

The paper presented a systematic method to validate the ML model's traffic state estimations and identify any inconsistencies by benchmarking it against well-defined traffic behavior according to fundamental physical conservation laws. The conservation law provided a standardized numerical approach to simulate "ground truth" traffic states. The validation focused on deviations from the Conservation Law by testing the model, which was trained on $v_f$=25 meters/sec, across different free flow speed ($v_f$) environments. Physics loss values were calculated by comparing the model's predictions to conservation law simulations for all test $v_f$ environments. These physics loss values were then normalized to allow performance comparison of the model across the various test $v_f$ conditions it was not directly trained on. Figure 4 displays the normalized physics loss results from the different $v_f$ environment tests, demonstrating the model's generalized performance when exposed to conditions beyond its training data.

\begin{figure}[!ht]
  \centering
  \includegraphics[width=0.5\textwidth]{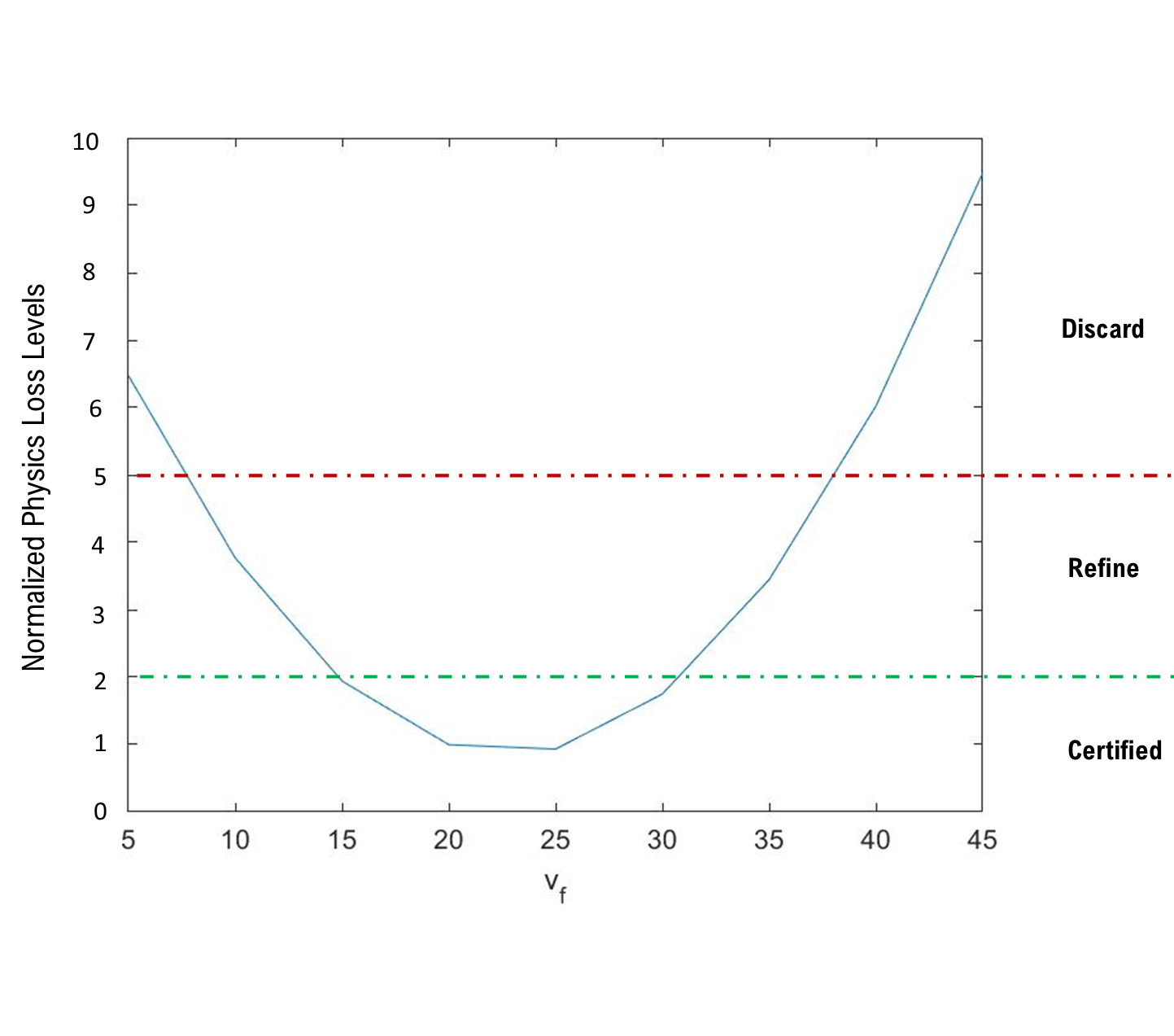}
  \caption{Classification based on Normalized Physics Loss.}\label{}
\end{figure}

Figure 5 clearly shows the model's Normalized Physics Loss (NPL) metric values at varying $v_f$ test environments from 5 to 45. NPL effectively measures the predictive error of the model in different conditions compared to its training environment of 25.
\begin{align*}
0< \text{NPL} \leq 2 &;& \text{Reuse} &:& C\\
2 < \text{NPL} \leq 5 &;& \text{Refine} &:& R\\
\text{NPL} > 5 &;& \text{Discard} &:& D
\end{align*}

\begin{figure}[!ht]
  \centering
  \includegraphics[width=0.5\textwidth]{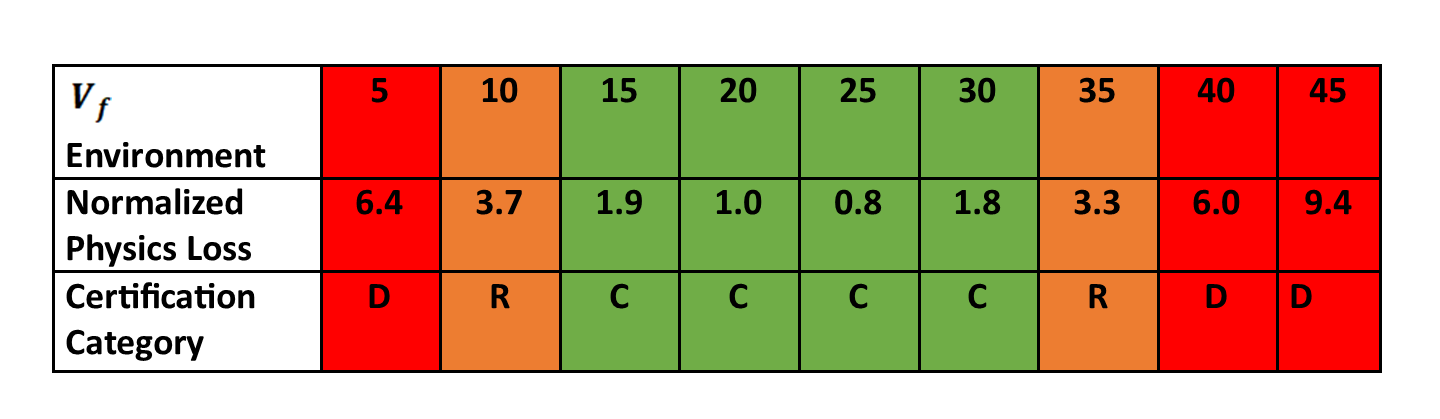}
  \caption{Certification classification table }\label{}
\end{figure}

The certification category definitions utilize NPL to objectively classify a model as reuse, refine or discard based on its robustness to domain shift. Models with low NPL (0-2) show acceptable performance for direct reuse, while those with intermediate NPL (2-5) may be refined with additional data or introduction of physics informed correction. Models with high NPL ($>$5) face significant domain shift and are discarded.

By applying these science-based metrics and definitions, the presented approach offers a standardized, reproducible method to assess and certify AI systems for safety and reliability when transferred to environments beyond their training scope. This has tangible applications, for example in certifying traffic prediction models across a range of real-world road conditions.

The table and category definitions demonstrate how the proposed methodology can guide careful model selection and development efforts to ensure models perform robustly, benefiting deployment in critical domains like autonomous vehicles. Overall, the paper presents an effective framework for certifying AI according to scientific rather than just economic factors.

\section{Summary and Conclusions}

This paper divulged a method of science-based certification of AI models to be applied in new operational environments. We investigated the viability and efficacy of certifying deep learning models for safety-critical traffic state estimation using physics rules. Enhancing deep learning models' dependability and credibility in applications where precise traffic state estimation is crucial for assuring safety was the goal and data is limited. 

The findings of this work pave the door for additional approaches such as physics-regulated deep learning model and transfer learning to better employ the models in new operating conditions. Furthermore, in this paper we used only synthetic data, future work should utilize the real world data for betterment of certification process of the AI models.

\section*{Acknowledgments}
This work is supported in part by the National Science Foundation (NSF) under Grant No. 2130990. 

© 2024 IEEE. Personal use of this material is permitted. Permission from IEEE must be obtained for all other uses, in any current or future media, including reprinting/republishing this material for advertising or promotional purposes, creating new collective works, for resale or redistribution to servers or lists, or reuse of any copyrighted component of this work in other works.


\bibliographystyle{ieeetr}
\bibliography{references}

\end{document}